\documentclass{article}
\usepackage{graphicx}
\graphicspath{{figs/}}

\usepackage{arxiv}

\usepackage[utf8]{inputenc} 
\usepackage[T1]{fontenc}    
\usepackage{hyperref}       
\usepackage{url}            
\usepackage{booktabs}       
\usepackage{amsfonts}       
\usepackage{nicefrac}       
\usepackage{microtype}      
\usepackage{lipsum}

\title{Multi-stage Clarification in Conversational AI: The case of Question-Answering Dialogue Systems}

\author{
  Hadrien Lautraite\\
  National Bank Of Canada\\
  600 de la Gauchetière\\
  Montréal, Québec\\
  \texttt{hadrien.lautraite@bnc.ca} \\
   \And
 Nada Naji \\
  National Bank Of Canada\\
  600 de la Gauchetière\\
  Montréal, Québec\\
  \texttt{nada.aj.naji@gmail.com} \\
  \And
 Louis Marceau \\
  National Bank Of Canada\\
  600 de la Gauchetière\\
  Montréal, Québec\\
  \texttt{louis.marceau@bnc.ca} \\
  \And
 Marc Queudot \\
  National Bank Of Canada\\
  600 de la Gauchetière\\
  Montréal, Québec\\
  \texttt{marc.queudot@bnc.ca} \\
  \And
 Eric Charton  \\
  National Bank Of Canada\\
  600 de la Gauchetière\\
  Montréal, Québec\\
  \texttt{eric.charton@bnc.ca} \\
}

\begin{document}
\maketitle

\begin{abstract}
Clarification resolution plays an important role in various information retrieval tasks such as interactive question answering and conversational search. In such context, the user often formulates their information needs as short and ambiguous queries, some popular search interfaces then prompt the user to confirm her intent (e.g. "Did you mean ... ?") or to rephrase if needed. When it comes to dialogue systems, having fluid user-bot exchanges is key to good user experience. In the absence of such clarification mechanism, one of the following responses is given to the user: 1) A direct answer, which can potentially be non-relevant if the intent was not clear, 2) a generic fallback message informing the user that the retrieval tool is incapable of handling the query. Both scenarios might raise frustration and degrade the user experience. To this end, we propose a multi-stage clarification mechanism for prompting clarification and query selection in the context of a question answering dialogue system. We show that our proposed mechanism improves the overall user experience and outperforms competitive baselines with two datasets, namely the public in-scope out-of-scope dataset and a commercial dataset based on real user logs.
\end{abstract}
\vspace*{0.2in}

\section{Introduction}
\label{intro}


Dialogue systems have been increasingly prevalent in many industries with the rise of virtual assistants such as Apple Siri, Amazon Alexa, and Microsoft Cortana. Such conversational agents can perform a variety of tasks such as, making transactions, booking appointments, or answering users' questions, among others \cite{ahmad2018review}.

In the context of question-answering agents, we often talk about \emph{intents}, which represent the various information needs or possible questions that users might have. Intents act as classes within the bot Natural Language Understanding (NLU) model. Such a model attempts at associating an incoming user message to one of the predefined intents learned from training data. Upon detection of an intent by the NLU model, the dialogue system will take a corresponding action, specifically, responding to the user with the answer associated with the detected intent. 
Under the hood, each intent is represented with several possible formulations in the training data since users can express themselves in a various ways to convey the same thought. As a concrete example, "forgot my password, what to do?" and "how to recover my pass-code" both relay the same intent but are phrased differently. This added complexity means that the intent classifier has to be generalized enough to handle unseen formulations. Since user queries are often short and ambiguous, the model might assign the wrong intent which yields an incorrect answer being given to the user. To address this issue, bots can have a clarification mechanism which engage the user to confirm or clarify their intent. 

In this paper, we focus on question-answering dialogue systems. Such agents can act as an additional communication channel that allows clients to ask a variety of questions and could therefore alleviate the pressure on the corporate call center for customer service and assistance. Additionally, dialogue systems constitute a more convenient tool than having to go through multiple possible answers returned by several searches on a traditional search engine. We propose a multi-stage clarification framework that allows to confirm the user intent before answering if the system's confidence is low and a mechanism to suggest some related formulations in case of the user's confirmation being negative. Evaluation on both click data from real interaction logs and human labeled data demonstrates the high quality of the proposed method, outperforming threshold optimization strategies.

The rest of the paper is arranged as follows: the next section outlines related work. Section \ref{sec:datasets} describes the datasets we used followed by the experimental setup in Section \ref{sec:experiments}. Afterwards, we present and discuss the results of our work in Section \ref{sec:results}. Finally, Section \ref{sec:conclusion} concludes our work and discusses future avenues. 

\section{Related Work}
\label{sec:relatedwork}



Interpretation issues are often the number one recurrent reason of bad user experience in dialogue systems \cite{folstad2020users}. Such issues translate as incapability of the dialogue system to understand the user request. In order to alleviate such issues, the dialogue system could trigger a fallback mechanism, that is, by answering that it does not understand the query or does not know the answer. Følstad and Brandtzaeg. \cite{folstad2020users} reports this behavior as the second most source of user dissatisfaction. The clarification process has been studied in various forms. In 1980, McKeown \cite{mckeown1980paraphrasing} presents a natural language interface to search in a database. The rule base system generate paraphrases to clarify the user intent in order to generate a database query that answer the user's needs.

Users tend to write short queries that are often ambiguous. This makes it challenging for a search engine or a dialogue system to predict possible intents, only one of which may pertain to the user query at hand\cite{zamani2020generating}. 
Search engines often use diversification to address this issue, by conveying multiple possible intents. Alternatively, the user is asked a question to \emph{clarify} her information need. This latter approach is essential for what is often referred to as “limited bandwidth” interfaces \cite{croft_talk_2019}, such as speech-only and small-screen devices\cite{zamani2020generating} yet is also found to be beneficial in web search \cite{belkin1995cases}. Such bidirectional interaction lends itself to dialogue systems.

Braslavski et al. \cite{10.1145/3020165.3022149} studied the different forms of clarification questions asked by humans on online forums such as the community question answering platform, Stack Exchange. The authors classify those questions in different categories including: requests for more information and questions in the form of "have you tried ..." .  Recent advances in the field of deep learning offer new possibilities in conversational AI.  Several studies propose neural networks architectures to rank possible responses in an information retrieval system \cite{crof2020IART}, \cite{10.1145/3340531.3412137}, \cite{guo2020deep}. Yang et al. \cite{crof2020IART} suggest categorizing user intents in forum discussions in classes, namely, original question, clarification questions, feedback or positive answer. The authors propose a new model named Intent-Aware Ranking with Transformers (IART) based on transformers \cite{NIPS2017_3f5ee243} in order to detect user intents and use it as an attention mechanism for ranking possible answers in a dialogue flow. Their proposed method leverages context when to decide whether to ask the user for further information or to provide an answer based on previous interactions. 
Zamani et al. \cite{zamani2020analyzing} developed a transformers-based model for ranking or selecting possible clarification question. Other studies \cite{zamani2020generating}, \cite{rao2019answer}, directly tackle the task of generating clarification questions. Rao and Daumé III \cite{rao2019answer} proposed an adversarial approach to generate clarification question. 

Asking too many clarification questions comes with a risk of deteriorating user experience due to overly inquisitive behavior. Sekulić et al. \cite{sekulic2021user} studied user engagement with the clarification pane in search engines in order to determine when and how to prompt users for clarification. Peixeiro et al. \cite{direct_answer2021} address the issue from an optimization perspective in order to determine the ideal threshold to maximize the number of correct direct answers for a maximum number of intents which in turn minimizes the number of unnecessary clarification questions.

We propose a simple yet effective method to provide users with the information they need while keeping a balance of direct answers and request for clarification. Instead of using question generation based on real interactions, which could expose us to data leakage \cite{carlini2020extracting}, we use canonical formulations from intents with similar keywords as clarification questions. Moreover, our proposed method does not require an additional ranking model to sort all possible clarification reformulations but rather use the confidence score from the initial natural language understanding module in order to rank the candidates canonical formulations. We show that our method improves effectiveness and allows for more fluid interactions. Its simplicity and the fact that no additional data are needed to train and maintain an supplementary ranking model makes our solution easy to deploy in real industrial context.

\section{Datasets}
\label{sec:datasets}
We conducted our experiments on two datasets. The \textbf{first dataset} is based on logs of real user interactions with our in-house corporate dialogue system. The dialogue system is deployed on our corporate transactional web platform. The dataset contains 8768 conversations collected during the first week of November 2020covering 272 distinct intents. We refer to this dataset as HOUSE.
The labels are inferred based on user interactions with the dialogue system. That is, when an intent is recognized by the NLU, the dialogue systems confirms the intent with the user "I understand you want to talk about ...", and if  the user clicks "yes" then an association is logged between the query and the intent. Such associations are used as ground-truth labels in our experiments. The dataset is mainly in French.


The \textbf{second dataset} is a publicly-available one known as the in-scope and out-of scope dataset designed by Larson et al.  \cite{larson2019evaluation} to train dialogue systems and evaluate their performance levels on a mix of \emph{in-scope} and \emph{out-of-scope} queries. In-scope queries can be mapped to an intent that is already known by the dialogue system (i.e., appears in the training set). Whereas an out-of-scope query represents a new or unknown concept to the dialogue system. For the purpose of our study, we use only the in-scope portion. The intents cover a variety of topics such as travel, banking, and car maintenance among others. Table \ref{tabl:in_scope_ex} presents some of the intents with some corresponding training examples. We refer to this datasett as SCOPE. The SCOPE dataset contains training and testing sets. We use the training set of 150 intents with 100 formulations each to train a dialogue system. As the evaluation of the dialogue system's performance with our clarification pipeline requires manual interactions, we focus our testing on the first 30 intents, considering only the first 10 formulations out of 30. The remaining 20  formulations are used as a validation set in order to fine-tune the fallback threshold of the dialogue system used as benchmark. 

\begin{table}[h]
\centering
\begin{tabular}{|l|l|}
\hline
\multicolumn{1}{|c|}{\textbf{Intent}} & \multicolumn{1}{c|}{\textbf{Examples}}                                                                                                                                             \\ \hline
translate                             & \begin{tabular}[c]{@{}l@{}}what expression would i use to say i love you if i were an italian\\ can you tell me how to say 'i do not speak much spanish', in spanish\end{tabular} \\ \hline
transfer                              & \begin{tabular}[c]{@{}l@{}}i need \$20000 transferred from my savings to my checking\\ complete a transaction from savings to checking of \$20000\end{tabular}                      \\ \hline
travel alert                          & \begin{tabular}[c]{@{}l@{}}does ireland have any travel alerts i should be aware of\\ does north korea have any travel alerts i should be aware of\end{tabular}                   \\ \hline
PTO request                           & \begin{tabular}[c]{@{}l@{}}how do i put in a pto request for the first to the ninth\\ am i allowed to put in a pto request for now to april\end{tabular}                          \\ \hline
oil change how                        & \begin{tabular}[c]{@{}l@{}}how do i change a car's oil\\ can you find instructions on how to change oil in a car\end{tabular}                                                     \\ \hline
\end{tabular}
\caption{Examples of intents and training samples from the SCOPE dataset}
\label{tabl:in_scope_ex}
\end{table}


\section{Methodology}
\label{sec:experiments}

Our dialogue systems are based on the Rasa Open Source framework. The pipeline consists of the following components: 
Firstly, a pre-processor which performs several NLP steps such as tokenization and featurization of the queries to obtain sparse representations at both word and character levels. The second component is Rasa's own intent classifier DIET \cite{bunk2020diet} with an NLU model that we trained for 200 epochs.  

During data preparation, we created a canonical formulation (one sentence) for each intent. This formulation describes the intent in natural language and is displayed to the user in the clarification pipeline to validate what she meant. For instance, "I understand that you want to talk about opening a new account, is that correct?" is the canonical formulation that is attached to the intent \emph{open new account}.

Our proposed multi-stage clarification pipeline encompasses the following stages:

\noindent \textbf{Stage 0 - Direct Answer}: in this stage, the dialogue system model \emph{understood} the user intent, that is the confidence level of the predicted intent is above the 75\% threshold. A direct affirmative response is given to the user.

\noindent \textbf{Stage 1 - Confirmation}: the dialogue system enters this stage when it is not sure to have understood, that is, the confidence level of the prediction is less than the threshold. Here, the dialogue system displays the canonical formulation of the predicted intent and asks the user whether it is a correct understanding or not. If the user answers "yes", the response attached to the detected intent is given. If the answer is "no", the system enters the next stage,

\noindent \textbf{Stage 2 - Suggestions}: here the dialogue systems displays several suggestions (up to six in our deployed systems) based on keywords appearing in the user query. The user can either choose one of the suggested canonical forms of the possible intents, otherwise can choose "none of the above". I the former case, the related response is given, in the latter, the dialogue system enters Stage 3. 

The keywords represent topics of interest related to intents. These topics typically represent products and services such as credit card, saving account, e-transfer, among many others. During training, each keyword, or combination of keywords, is linked to intents whose canonical formulations contain at least one of those keyword. These intents' canonical formulations represent the suggestions in this stage.

\noindent \textbf{Stage 3 - Frequently Asked Question (FAQ)}: this is the last stage in the pipeline which provides the user with general, query-independent, recurrent questions and answers. This serves as a fallback procedure that offers helpful questions in a dynamic manner as it allows the user to navigate topics in breadth and in depth.

\begin{figure}
\centering
\caption{Our proposed multi-stage clarification framework with examples showing the four distinct stages: Stage 0) Direct answer, Stage 1) Confirmation, Stage 2) Suggestions and, finally, Stage 3) General FAQ.}
\label{fig:direct_answer}

\begin{tabular}[t]{ll}
    \begin{tabular}[h]{l}
        \fbox{\includegraphics[width=5.5cm]{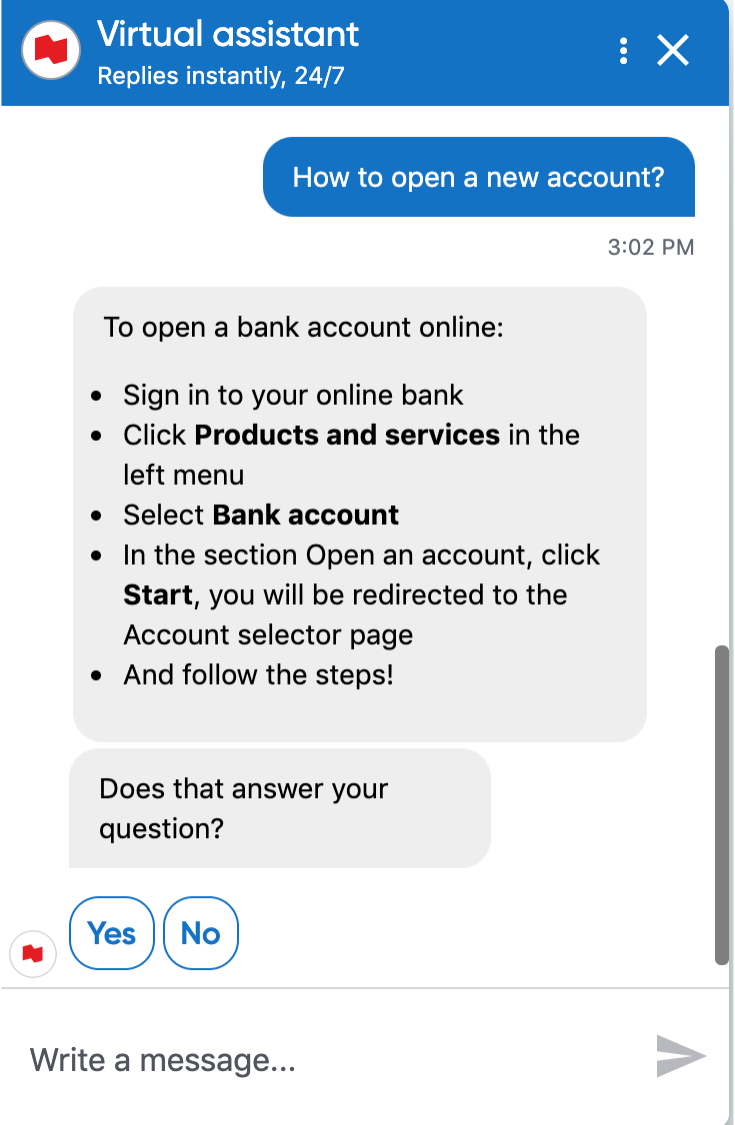}} \\ Stage 0) Dialogue system giving a direct answer \\explaining how to open an account\\ \\ \\ \fbox{\includegraphics[width=6cm]{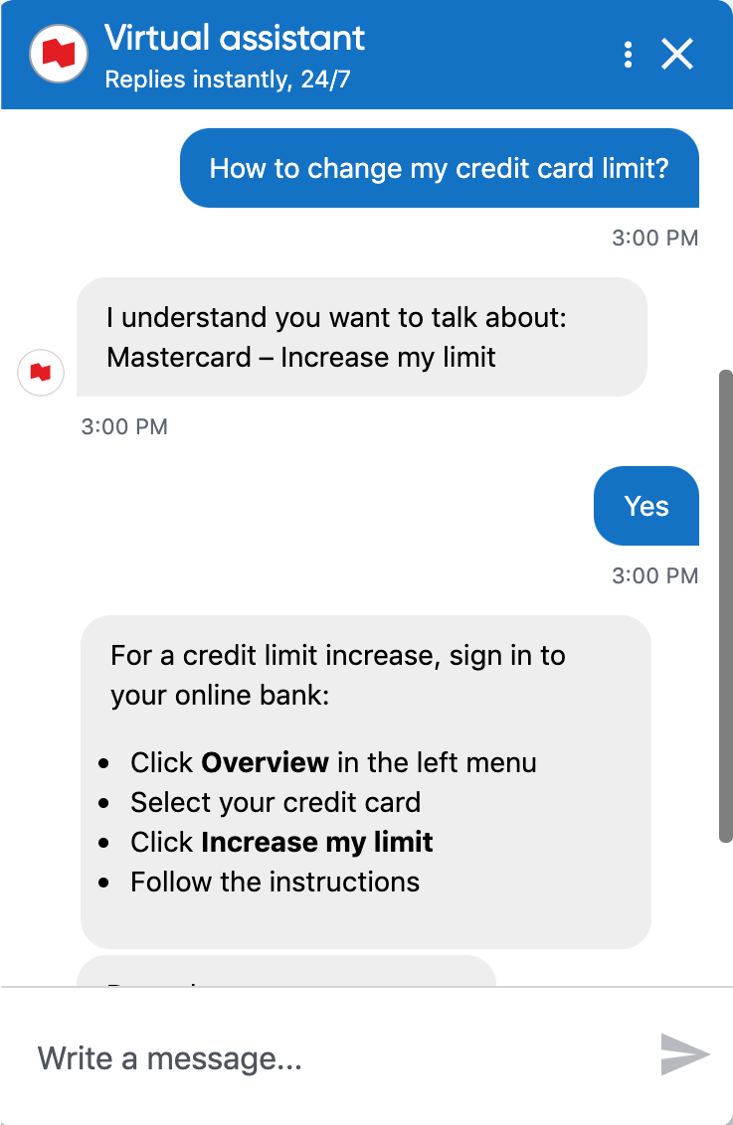}} \\ Stage 1) Dialogue system giving an answer \\during the Confirmation stage
        \end{tabular}
    &
    \begin{tabular}{l}
    \fbox{\includegraphics[width=5.5cm]{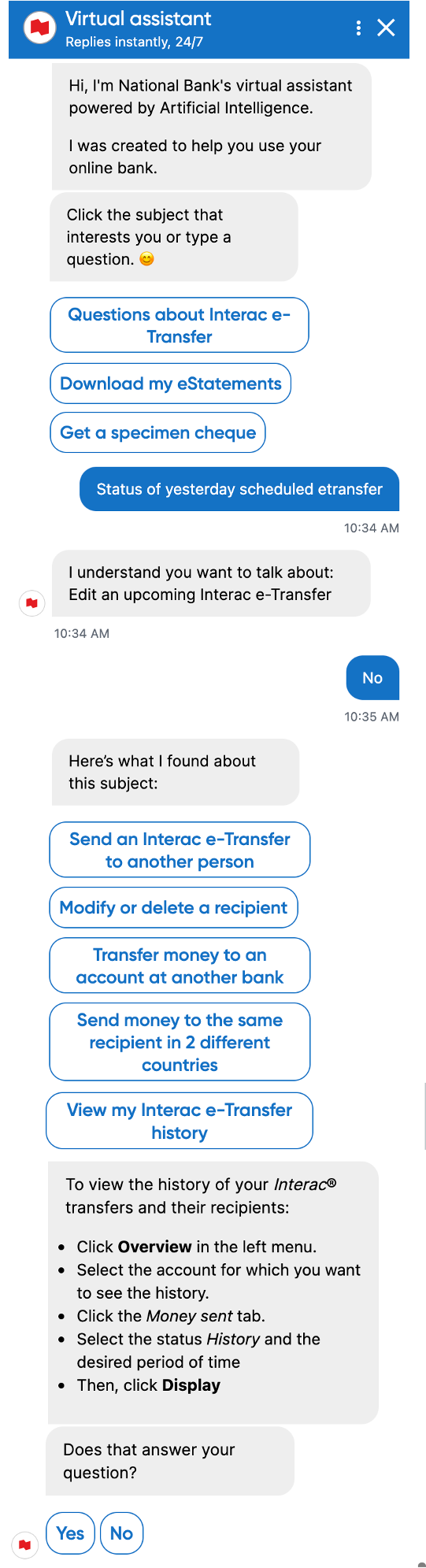}} \\ Stage 2) Dialogue system giving an answer \\during Suggestions stage. The user chose the \\5th suggestion "View my Interac e-Transfer \\history" and got the corresponding response
    \end{tabular}
    

\end{tabular}
\end{figure}

Figure \ref{fig:direct_answer} illustrates the different stages of our dialogue system with examples from the HOUSE corpus in English.


\begin{figure}[h]
\centering
\begin{tabular}[t]{ll}
  \fbox{\includegraphics[width=5.5cm]{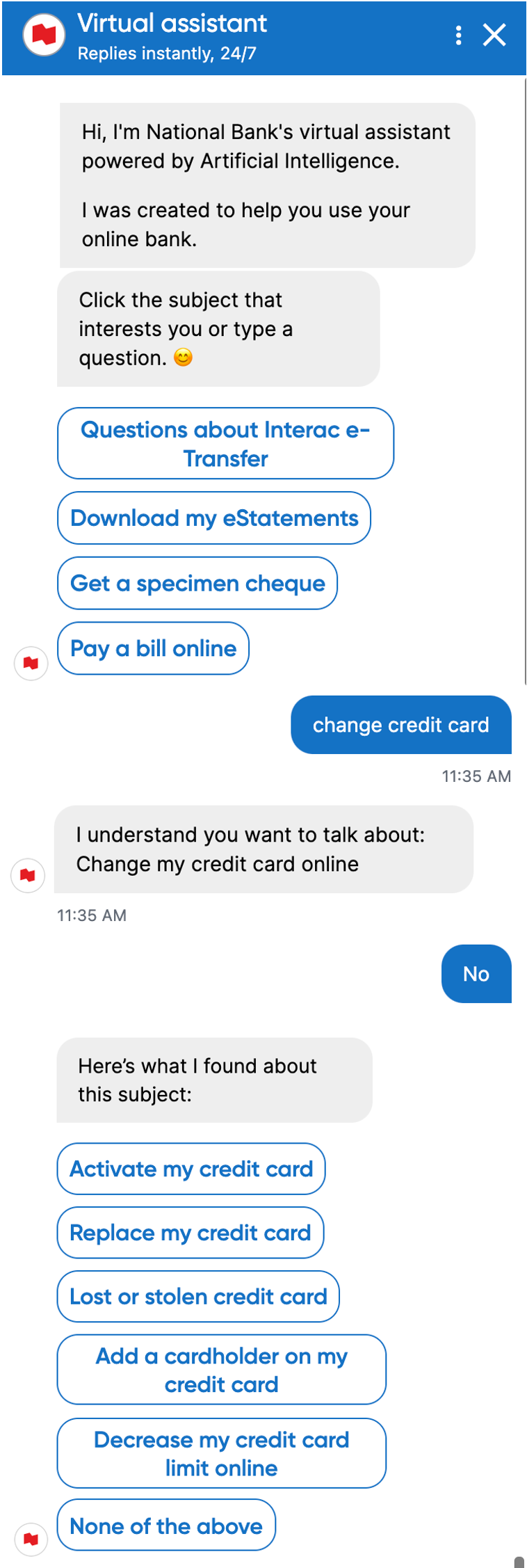}}
    &
    \fbox{\includegraphics[width=5.5cm]{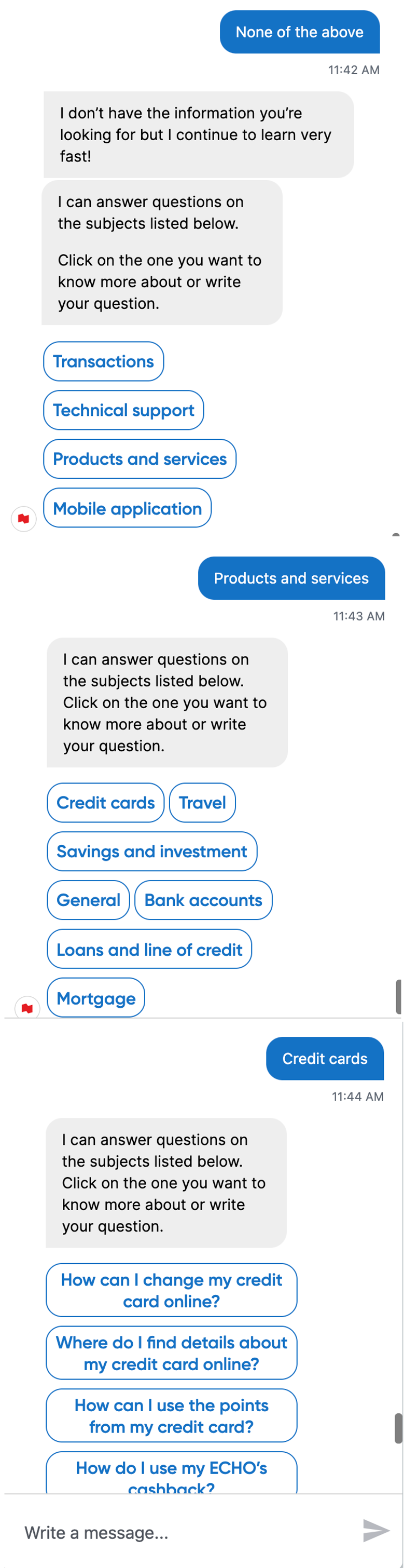}}
\end{tabular}
\\
Stage 3) None of the suggestions satisfactory, generic topics (FAQ) provided
\end{figure}

\clearpage
\section{Results and Discussion}
\label{sec:results}
In this section, we present our results and observations. We begin by analyzing the results obtained from the HOUSE dataset:

The dialogue system directly answers the user's question in more than 40\% of the interactions. Roughly half of the interactions entered later stages of the clarification pipeline. The remaining 10\% is composed of aborted conversations or the user jumping to another question rather than acting on the confirmation or suggestions.

The Confirmation stage allows the dialogue system to give an answer to the the user (positive confirmation) for 29\% of the discussions that go through the clarification process. This 10\% of the total number of interactions would not have been responded in a dialogue system in the absence of the multi-stage clarification process and would have ended with a fallback mechanism.
Asking for confirmation allows to answer the client's needs without taking the risk of harming the user experience by giving a wrong answer. 

\begin{figure}[h]
    \centering
    \includegraphics[width=10cm]{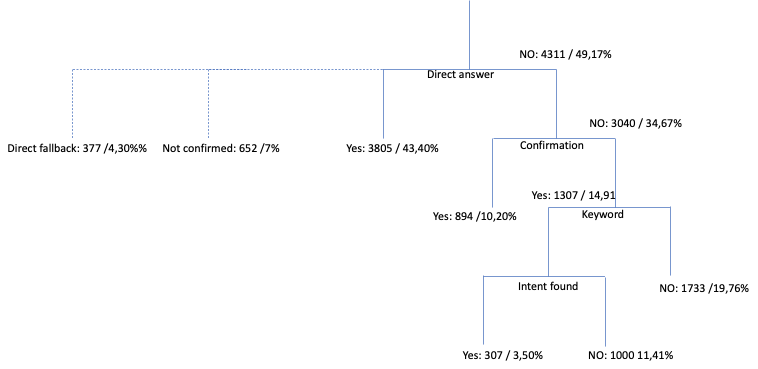}
    \caption{Overall interactions volumes and percentages among the various step of our dialogue system}
    \centering
    \label{fig:bot_tree_interactions}
\end{figure}

In case of a negative answer in the Confirmation stage, the dialogue system will enter the Suggestions stage to propose to the user several possible intents. This mechanism allows to identify the correct intent in 3,5\% of the total interactions or 10\% of the interactions that go through the clarification process.

figure \ref{fig:bot_tree_interactions} depicts the distribution of user interactions with the dialogue system across the various stages of our dialogue system based on the HOUSE dataset.

Our experiments show that our proposed clarification framework allows:
\begin{enumerate}
\item To give a correct answer to the user in 10\% of the interactions even if the confidence was lower than the threshold. Those 10\% would have ended with a fallback mechanism in a classical dialogue system
\item To give the correct answer to the user when the predicted intent is wrong in 3,5\% of our experiments by suggesting pertinent canonical formulations of the user query.
\end{enumerate}

Since the data is not annotated, we do not study the impact of direct answer threshold optimization. However a quick analysis shows that two third of the intent predictions where the model confidence was just bellow the threshold (threshold - 0.1) were not confirmed by the user during the clarification stage of our clarification pipeline. We conduct proper comparison with threshold optimisation in the second experiment.

In the second experiment, on SCOPE, we compare the performance of two dialogue systems. One with our proposed clarification pipeline and one with a simple fallback procedure. With the fallback mechanism, the dialogue system does not give answer if its confidence is below a certain threshold. Both dialogue systems have been trained using Rasa DIET classifier with 100 epochs and they share the same NLU engine. \par
Our clarification pipeline uses keywords to suggest new answers if the user responds negatively to the validation stage. In order to find those keywords we perform a TFIDF analysis on the training examples. Then, we select the top five words with highest TFIDF per intent as keywords for our clarification pipeline. \par
Finally, our vanilla dialogue system can benefit from the use of threshold optimization. We optimize to find the best possible threshold in order to maximize the number of correctly answered questions. In order to do so we look at the dialogue system performance on the validation set. We select the threshold that maximize the number of good responses to the user queries.\par

We start by comparing the performance using 0.75 as fallback threshold for the simple vanilla dialogue system. For our improved dialog engine, we also use 0.75 as threshold to trigger the clarification pipeline and 0.3 for direct fallback.

The dialogue system with the clarification component can answer directly 257 (86\%) of the user queries with an accuracy for direct answers of 94\%. In total 80\% of the queries get a correct direct answer. For 3 queries (1\%) the dialogue system was confused and couldn't answer the user question (confidence < 0.3). At the CONFIRMATION stage, 45\% (18) of the queries have a confirmed intent. For the remaining 22 questions, the dialogue system is able to propose alternative answers through SUGGESTION in 95\% (21/22) of the cases. The correct answer is among those propositions in 90\% (19/21) of the cases. In total our dialogue system is able to answer the client needs for 93\% (277/300) of the interactions, with only 5\% (16/300) of wrong answers and 2\% (7/300) of fallback. 

The dialogue system with simple fallback mechanism and a non optimized threshold set to 0.75 give a correct answer in 80\% (241/300) of the interactions, a wrong direct answer in 4\% of the cases and the conversation ends with a fallback in 14\% of the interactions. 

We select the fallback threshold that maximize the number of correct answers given to the user. The selected threshold is 0.35. With such a threshold, the dialogue system gets the following performances: good answers: 86\% (259/300), bad answers: 13\% (38/300) and fallback: 1\% (3/300).

Finally we compare the models performances in term of F1 score: the harmonic mean between precision and recall. The dialogue system with a non-optimized threshold and the dialogue system with our disambiguation component get the same macro-F1 score. By lowering the threshold, the precision of the dialogue system with optimized threshold decrease which leads to lower results in term of macro-F1 score. Regarding the micro-F1, our proposed method get the best results. 

\begin{table}[h]
\centering
\begin{tabular}{|l|l|l|l|}
\hline
\textbf{}    & Simple fallback mechanism & Optimized fallback mechanism & clarification mechanism \\ \hline
Good answers & 80.3\%                   & 86\%                         & 92.3\%                 \\ \hline
Bad answers  & 5.3\%                    & 13\%                         & 5.3\%                  \\ \hline
Fallback     & 14.3\%                   & 1\%                          & 2.3\%                  \\ \hline
macro-F1     & 0.64                      & 0.51                         & 0.64                    \\ \hline
micro-F1     & 0.8                       & 0.86                         & 0.92                    \\ \hline

\end{tabular}
\caption{performances comparison between three dialogue systems: with a simple fallback mechanism, fallback mechanism with optimized threshold and our proposed clarification component}
\label{tabl:results2}
\end{table}

Our clarification pipeline allows us to increase the performance in answering the client's needs by 15\% with regard to the vanilla dialogue system and 7\% compared to the dialogue system with optimized threshold. The later suffers an increase in the number of incorrect answers whereas our dialogue system can achieve this performance without increasing the number of bad responses.

\begin{figure}[h]
    \includegraphics[width=6.5cm]{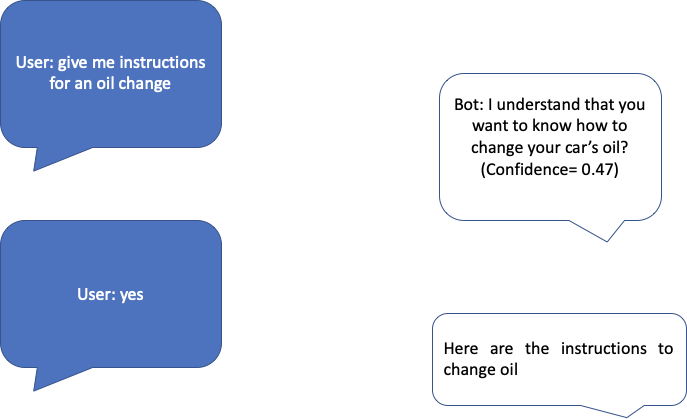}
    \hfill
    \includegraphics[width=6.5cm]{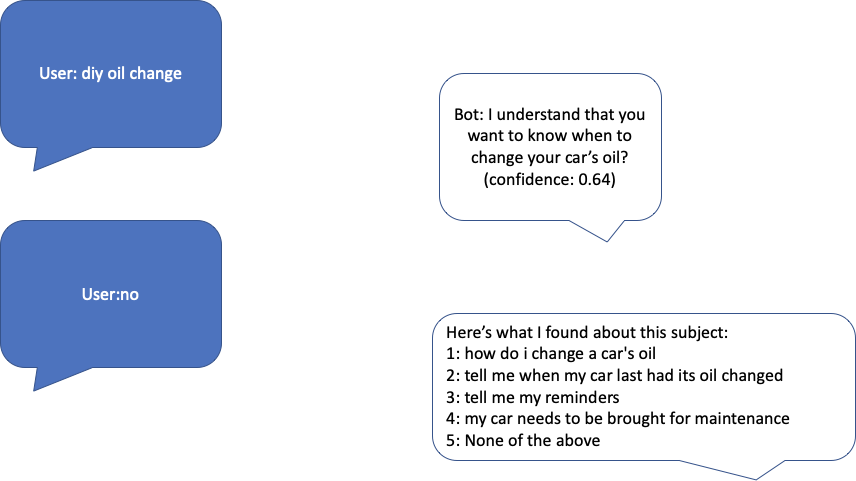}
    \caption{Example of our clarification pipeline for two relatively close intents}
    \label{fig:desambigu}
\end{figure}

Figure \ref{fig:desambigu} illustrates our proposed clarification pipeline in a situation where two intents are very close semantically . We believe our proposed method can prevent the user from the frustration of not being able to find the precise answer to his questions but rather being redirected to something similar. 

\section{Conclusions and Future work}
\label{sec:conclusion}
In paper we propose a multi-stage clarification framework. We show that our proposed framework improves the performance of the dialogue systems. This in turn improves the user experience as relevant answers are given and clarification is triggered only when needed. This framework reduces the risk of providing hasty, inaccurate answers to the user. When unsure of the user's intent, the dialogue system prompts for confirmation or suggests possible formulations without being unnecessarily highly inquisitive. 
Our method is simpler than related work on clarification question generation and ranking and is relatively straightforward to deploy and monitor without the need of extra data or model. We conducted our evaluations on two datasets. On the publicly-available in-scope out-of-scope \cite{larson2019evaluation} dataset our proposed clarification pipeline allow us to increase the performance in answering the client's needs by 15\% with regard to a baseline dialogue system. 
As a future direction, we will explore click bias and patterns on the interaction with the dialogue system, how the results might differ by device, conversation length/stage, and order of the suggestions. Further work may also include further customer-specific answers and clarification questions based on click behaviour and implicit feedback or using external info held on the client (bank account, previous transactions) to propose better answers and clarification.

\bibliographystyle{unsrt}  
\bibliography{references}
\end{document}